# PARALLEL WiSARD OBJECT TRACKER: A RAM-BASED TRACKING SYSTEM

Rodrigo da Silva Moreira[1] and Nelson Francisco Favilla Ebecken[2]

[1]Department of Civil Engineering, UFRJ University, Rio de Janeiro, Brazil
[2]Department of Civil Engineering, UFRJ University, Rio de Janeiro, Brazil

## ABSTRACT

*This paper proposes the Parallel WiSARD Object Tracker (PWOT), a new object tracker based on the WiSARD weightless neural network that is robust against quantization errors. Object tracking in video is an important and challenging task in many applications. Difficulties can arise due to weather conditions, target trajectory and appearance, occlusions, lighting conditions and noise. Tracking is a high-level application and requires the object location frame by frame in real time. This paper proposes a fast hybrid image segmentation (threshold and edge detection) in YcbCr color model and a parallel RAM based discriminator that improves efficiency when quantization errors occur. The original WiSARD training algorithm was changed to allow the tracking.*

## KEYWORDS

*Ram memory, WiSARD Weightless Neural Network, Object Tracking, Quantization*

## 1. INTRODUCTION

Surface targets (ships or enemy vessels) tracking is a task of great importance for warships. Advances in electronics and the emergence of new technologies enabled the design and implementation of modern weapons. Such weapons can be supported by a tracking system that calculates the best direction for a shot.

Video tracking is a tool that replaces or assists radar tracking systems. Many factors difficult radar tracking, including clusters, sectors not covered by radar, electronic countermeasures systems and high cost.

This paper presents an efficient video target tracker based on the WiSARD weightless neural network, which is able to work in real time and can compensate quantization errors generated by the image segmentation. Quantization in this paper means the conversion of pixel values to a binary representation. The WiSARD neural model is based on RAM-based neurons and various RAM nodes sizes in networks working individually and in parallel were tested together with image segmentation methods and color models. Each target is searched only in a small region called the region of interest (ROI) in order to reduce execution/processing time.

Tracking is generally a challenging problem [1] and has been widely studied. It is used for surveillance [2], target tracking, human gestures recognition, monitoring of traffic on the streets, and even to help a robot to do inertial navigation. Trackers were developed based on background removal [3], feature points [4], feed-forward neural networks [5], fuzzy logic [6] and others methods. Trackers based on weightless neural networks (WNN) are not widespread or doesn't exists. This article introduces PWOT, an object tracking system based on the WiSARD WNN.

The complexity of tracking sea surface targets in videos comes from several factors such as climate and sea conditions, target speed and appearance and presence of non-hostile vessels. The proposed Parallel WiSARD Object Tracker (PWOT) [7] (figure 1) has tree components: an object detector ObjDet based on the WiSARD WNN (first RAM memory), a second RAM memory (RAM2) and a position predictor. At each frame, the ObjDet returns the target position and writes it into RAM2. The WiSARD has a group of discriminators. Each one acts as a classifier, recognizing a different class of bit pattern. At the first frame, all discriminators are trained with the quantized pixels (target model) inside a frame region defined manually by the operator, the selection window (SLW). At the following frames, the ObjDet receives as input the quantization result of all pixels inside the ROI. Each discriminator tries to recognize the target in a different region inside ROI. Comparing the discriminators responses, the ObjDet defines the target position. The position predictor estimates the object position in the next frame. The ROI center will be moved to this position at next frame. It was developed in C++ and QT. The videos run on a HP Lap-Top with 298MB RAM memory, 1.14GB HD, AMD Turion processor and a Linux operating system. We used an outdated computer to test the effectiveness of this approach on a lagged device.

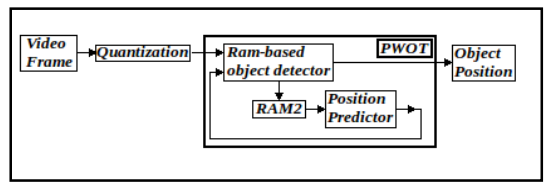

Figure 1. Parallel WiSARD object tracker

## 2. THE WEIGHTLESS NEURAL NETWORK WISARD

WNN is an important branch of research related to neural networks. The neuron input and output are sets of bits and there are no weights between neurons. The activation function of each neuron is stored in look-up tables that can be implemented as RAM memories [8]. The training of an artificial neural network (ANN) involves the adjustment of weights. Unlike these networks, the WNN training process is carried by modifying the words stored in the look-up tables, allowing the construction of flexible and fast training algorithms. With the use of look-up tables, any activation function can be implemented at the nodes, since any binary value can be stored in response to any set of input bits during training. The fast training speed is due to the mutual independence between nodes when a new input data is presented. The training process of an ANN changes the weights values and the network behavior relative to patterns previously trained is modified.

The WiSARD neural network [9] is a WNN where neurons are implemented as RAM memories. The impossibility of implementing XOR function with the perceptron is a bypassed problem and the WiSARD can be trained in very short time, which is essential for tracking.

The WiSARD is designed for pattern recognition but can also be used for other purposes, such as target tracking (this paper), the automatic movement control of an offshore platform [10], pattern recognition [11], minimization of the neuron saturation problem [12], development of surveillance systems [2], inertial robot navigation [13] and the diagnosis of neuromuscular disorders [14]. The WiSARD can be easily implemented in hardware. The nodes can be easily implemented as RAM memories, and hence, the PWOT hardware implementation is straightforward.

The WiSARD is built grouping a set of basic elements called discriminators. Each discriminator is a set of k RAM memory nodes (figure 2), each addressed by N bits (N-input RAM node) designed to recognize a different class of bit pattern. Each RAM stores 2N words of one bit [8]. At an image containing k.N pixels, one quantized pixel represents one bit. k sets of N randomly chosen bits are connected to k RAM bus addresses. The discriminator response is a weightless sum of all k RAM accessed words. Binding of k.N bits at bus addresses is called input mapping. Once the input mapping is set, this remains constant during the training and classification phases.

Suppose that a WiSARD discriminator has k RAM nodes. Before the discriminator training phase, the bit "0" is written at all RAM accessed addresses. A training vector set X with class A examples, k.N bits size each, is prepared. During the training phase, each example from X is placed at the discriminator input, one by one. The bit "1" is written at all RAM accessed address of the discriminator being trained. Another training vector set Y with class B examples is used to train another discriminator. In the classification phase, one test vector is placed at the WiSARD input. The vector is classified as a class member represented by the discriminator that returns the greatest response. Each discriminator recognizes one different letter and is trained with quantized images containing only the letter it is designed to recognize.

The proposed tracker PWOT was tested on sea surface targets. At each video frame, pixels are quantized and placed at the ObjDet input (first PWOT component). ObjDet is based on the WiSARD. At each frame ObjDet try to find the target inside the ROI. The ObjDet discriminators are not trained in the same way as the original WiSARD training. To find the target position, all discriminators are trained with the same input bit pattern: the SLW quantized pixels. PWOT should return the target position. For this task, the WiSARD training procedure was modified. All discriminators are trained with the same target model (quantized SLW pixels). However, each discriminator tries to recognize the target at a different region (figure 3). The search region covered by each discriminator is spatially displaced from the others. The discriminator that returns the highest response with a confidence level C (relative difference from the second highest response) greater than a minimum value reveals the target position.

The search region size of one discriminator is equal to the SLW size. The union of all discriminators search regions forms the ROI. The search region of each discriminator has nonempty intersection with respect to their neighbors (Fig 4). The algorithm steps are:

Step1 (first frame): The operator selects manually the target in the first video frame to set the SLW [15];
Step2 (first frame): convert the pixels inside the SLW to a binary representation (pixel quantization);
Step3 (first frame): Train the WiSARD (each ObjDet discriminator) with the SLW quantized pixels;
Step4 (following frames): Track the target (trained bit pattern) with PWOT. Repeat step 4 at next frame;

The ROI geometric center is initially located at the SLW geometric center. The position predictor estimates the target position at next frame using a simple Kalman filter. Then, PWOT moves the ROI geometric center to the estimated position (step 4).

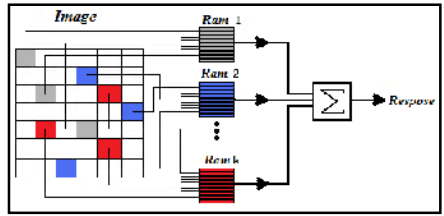

Figure 2. WiSARD discriminator: a discriminator with k RAMs of 9bit bus address where every bit is connected to a quantized pixel. Being represented only three of nine pixels connected to each RAM.

Step4 summary: ObjDet is a set of discriminators implemented as RAM memories (RAM1). At each frame, the ROI pixels are quantized before PWOT acts. The quantized pixels contained inside the search region of each discriminator are placed at their input as a binary test vector. All discriminators try to recognize the target inside their search region. The discriminator that returns the highest response with a minimum degree of confidence C reveals the target position (figure 3). ObjDet writes the target position at a FIFO stack (RAM2). Only the N lasts target positions are saved. The position predictor reads all the N target positions to estimate it. The ROI geometric center is moved to the estimated position. This way the ObjDet will search the target where it will probability be in the next frame. When PWOT perceives that the answers of two discriminators are similar (low value C), the WiSARD configuration must be changed online.

This algorithm can become automatic by substituting the first step with a moving detector and a classifier. The classifier should decide if the object inside one moving region is a vessel or not and return the target bounding box (SLW). Fast algorithms for detecting moving objects such as background subtraction [3] and frame difference [4], and for classifying objects based on WiSARD [14], ANN, SVM or Bayesian networks [1] are extensively used.

# 3. PWOT

This section addresses the simulations performed to improve the Parallel WiSARD Object Tracker efficiency.

## 3.1. Video characteristics

All videos have resolution of 640x480 pixels (figure 4). The scene contains only one surface target with not real and very difficult maneuvering.

## 3.2. Image Segmentation

The novel hybrid image segmentation method used in all simulations up to the discriminator setting GP6 (section 3.3.2) receives as input RGB images. The method segments the image by sampling, threshold and edge detection:

Step 1- The SLW center cs selected by the operator belongs to the target. Four points at the target border are calculated by sliding the Prewitts edge detector in transversal and longitudinal directions departing from cs (figure 5). The transverse (p1-p2) and the longitudinal bands (p3-p4) pf are good pixel samples to calculate the thresholds because ships usually have little tonal variations. To minimize the influence on the simulation, the initial SLW was kept constant for each video. The sampling provides a rapid image segmentation, essential for tracking. The standard segmentation methods are computationally more expensive.

Step 2- Calculate the mean med_pf and standard deviation s_pf of the pixels pf at R, G and B channels.

Step 3- Sample the pixels pw belonging to the sea and calculate the mean m_sea at R, G and B channels.

Step 4- Calculate (1) six thresholds in the first frame, a pair for each RGB channel. Each pair defines a threshold rule that includes the target pixels and exclude background pixels. x, y and z values are such that best separate pf from pw pixels by thresholding.

$$L1 = m\_pfR + x.s\_pfR; \quad L2 = m\_pfR - x.s\_pfR;$$
$$L3 = m\_pfG + x.s\_pfG; \quad L4 = m\_pfG - x.s\_pfG; \quad (1)$$
$$L5 = m\_pfB + x.s\_pfB; \quad L6 = m\_pfB - x.s\_pfB;$$

Step 5- Quantize the ROI pixels at each frame after the first using the six calculated thresholds.

This segmentation algorithm performes better than the following methods: threshold [16], edge detection [17], threshold with edge detection [18], region growth [16], Watershed Transformation [19] and Split and Merging [16].

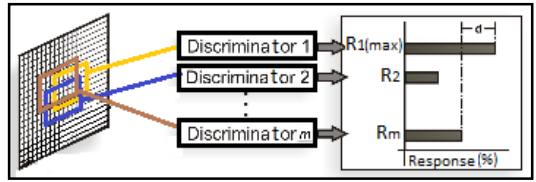

figure 3. The seach region of each discriminator has a different color (left). Responses of all discriminators (right).

Table 1. Videos

| N | SC | NF | SA |
|---|----|----|----|
| VF | F | 55 | 40x17 |
| VB$_1$ | B | 79 | 68x24 |
| VB$_2$ | B | 66 | 68x19 |
| VZ | Z | 87 | 42x38 |
| VC | C | 60 | 36x25 |
| VS | S | 50 | 61x21 |
| VP$_1$ | P | 79 | 40x21 |
| VP$_2$ | P | 79 | 47x21 |
| VUW$_1$ | UW | 75 | 34x29 |
| VUW$_2$ | UW | 28 | 55x26 |
| VAE | AE | 78 | 80x27 |

At all tables N means video name; NF: Number of frames; SC: Ship class; SA: Starting selection region size (pixels).

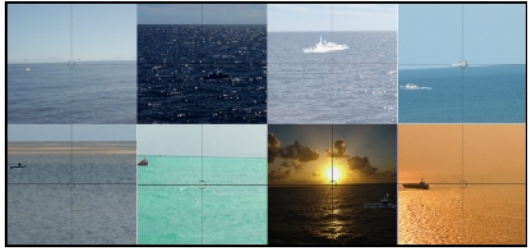

Figure 4. The videos VF, VB1 (and VB2), VZ, VC, VS, VP1 (and VP2), VUW1 (and VUW2) and VAE are disposed from left to right, from top to bottom.

## 3.3. Experiments

All discriminators initially have 9-input RAM nodes (section 3.3.1 and 3.3.2). Other RAM sizes were tested next. The discriminators centers are displaced from each other xp pixels relative to x axis and yp pixels relative to y axis. The geometric center of each discriminator search region forms a grid of points GP. Tracking errors occur when the tracker indicates a position with less than 50% ratio of intersection/union with the right bounding box.

### 3.3.1 Setting the WiSARD discriminator

The S1 simulation (table 2) aims to choose the optimal discriminator quantity. The greater the quantity, greater ROI becomes and lower is the probability of target missing, but greater is the mean execution time ET of the tracking algorithm at one frame. However, at VF video the tracker missed the target 3 frames before with GP3 compared to GP1 and GP2 due to quantization errors. At VS, the target is lost in the third frame with GP1, GP2 and GP3. The target direction changes instantaneously. It comes out from the ROI. But with GP4 no target loss occurs because the search region size is bigger.

With a more accurate quantization, the tracker fails less. At VF, VC, VUW1 and VUW2, GP4 provides worse performance compared to GP1, GP2 and GP3 with respect to the frame number that the first error occurs despite ROI being bigger due to SLW pixel quantization errors (wrong target model). With GP5, the performance is inferior compared to GP4 at VF and VC for the same reason. The position predictor indicates a dubious region some frames earlier.

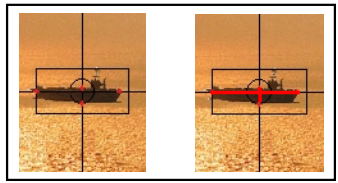

figure 5. (left) points p1, p2, p3 e p4 in SLW; (right) two pixel band samples.

The introduction of position predictor based on Kalman filter improves the tracker performance on most videos. However, with GP5 the tracker fails before compared to GP4 at VS due to the unpredictable target movement that tricks the predictor.

Considering the ET and the performance related to the frame that the first error occurs, the GP5 setting was more efficient proving the importance of including a position predictor in PWOT.

### 3.3.2 Improving the hybrid image binarization method proposed

Several image segmentation methods and color models were tested. The combination of color model YCbCr with a hybrid segmentation method achieved the best results, being used in all settings from GP7.

Table 2.  Simulation S1: Mean tracking time ET(ms) x Frame that occurred the first tracking error.

| N | $T_1$ | $T_2$ | $T_3$ | $T_4$ | $T_5$ | $M_1$ | $M_2$ | $M_3$ | $M_4$ | $M_5$ |
|---|---|---|---|---|---|---|---|---|---|---|
| VF | 50 | 101 | 139 | 83 | 82 | 35 | 35 | 32 | 3 | 3 |
| $VB_1$ | 102 | 224 | 351 | 233 | 242 | 42 | 64 | 70 | X | X |
| $VB_2$ | 105 | 203 | 318 | 214 | 220 | 63 | 70 | X | X | X |
| VZ | 121 | 269 | 427 | 253 | 267 | X | X | X | X | X |
| VC | 51 | 91 | 161 | 109 | 105 | 53 | X | X | 29 | 19 |
| VS | 85 | 165 | 240 | 176 | 190 | 3 | 3 | 3 | X | 58 |
| $VP_1$ | 123 | 254 | 391 | 250 | 257 | 34 | 71 | 73 | 77 | X |
| $VP_2$ | 155 | 333 | 493 | 352 | 355 | 32 | 70 | 73 | 78 | X |
| $VUW_1$ | 65 | 129 | 247 | 130 | 130 | 67 | X | X | 71 | X |
| $VUW_2$ | 82 | 181 | 252 | 159 | 184 | 17 | 24 | 26 | 13 | 23 |
| VAE | 104 | 239 | 412 | 239 | 242 | X | X | X | X | X |

T1, T2, T3, T4 and T5: ET (ms) with GP1, GP2, GP3, GP4 and GP5 settings respectively; M1, M2, M3, M4 and M5: number of the first frame that a tracking failure occurred with GP1, GP2, GP3, GP4 and GP5 settings respectively. GP1: 12 x 12 discriminators spaced 2 xp pixels (sxp=2) and 2 yp pixels (syp=2) in relation to its neighbors. GP2: 20 x 20 discriminators with sxp=2 and syp=2. GP3: 30 x 30 discriminators with sxp=2 and syp=2. GP4: 20 x 20 discriminators with sxp=5 and syp=5. GP5: 20 x 20 discriminators with sxp=5 and syp=5 and the tracker has a position predictor. At all tables X indicates that the tracker do not fail.

The improved image segmentation method algorithm has two steps:

Step 1- Calculate the pixels pf the same way as in step 1, section 3.2; transform from RGB to YCbCr model and calculate the mean and standard deviation of the pixels pf at Y, Cb and Cr channels;

Step 2- Calculate (2) six thresholds. At all videos, x=3, y=3 and z=1.5 are sufficient to eliminate tracking errors caused by an inefficient quantization (table 3). At the frames after the first, the ROI pixels are transformed to YcbCr model and quantized with these thresholds.

$$L1=m\_pfCr +x.s\_pfCr; \quad L2=m\_pfCr – x.s\_pfCr;$$
$$L3=m\_pfY +x.s\_pfY ; \quad L4=m\_pfG – x.s\_pfG; \quad (2)$$
$$L5=m\_pfCb+x.s\_pfCb; \quad L6=m\_pfCb – x.s\_pfCb;$$

Using GP5, the tracker loses the target before the end at VF, VC, VS and VUW2. To improve the performance, two settings, GP6 and GP7, which have two layers of discriminators, were tested (table 3). By hypothesis, increasing the discriminator density near the position indicated by the predictor the tracker fails less. The GP6 quantization method is the same as the used in section 3.2. The GP7 quantization method is an evolution of this method.

The introduction of a second discriminating layer, GPD, denser and smaller than GS, improved the tracking at VF, VT and VS, but at VUW2, using GP6 setting, the WiSARD response is greater than using GP5 on a sea surface region 9 frames earlier caused by quantization errors. VUW1, VUW2, VB1 and VB2 are very challenging because the target-sea contrast is low, producing more quantization errors. Using a more efficient and innovative quantization method, the GP7 setting allows a right target track until the end of all videos, proving the efficiency of this method for tracking sea surface targets and the quantization influence on the tracking performance.

### 3.3.3 Optimal RAM node size

Seeking to lower ET, a GP7 was used to check what RAM size reduces ET and maintains the same efficiency. Addressing RAM node size varying from 1 bit until the largest size allowed by the computer memory (figure 6). The graph shape is the same to all tests, changing only the node size where ET starts to increase very much (from 16 to 22 bits). ET increases as the computer RAM memory limit is exceeded.

Seeking to study how the RAM node size can compensate quantization errors and improve the tracker performance, the quantization quality was purposely made worse by various degrees, modifying the 6 thresholds values. All supported sizes were tested. Considering only ET, nodes with bus address size between 2 and 14 bits have similar performance, but with 3 bits the tracker compensates better the quantization errors, failing less in about 70% of the cases. It is concluded that the node size influences the robustness against quantization errors.

### 3.3.4 Quantity of Discriminators

VF, VC and VUW1 were used to verify if there is a relationship between density/number of discriminators and robustness against quantization errors. At these videos the quantization quality was purposely worsened. Four discriminator settings were tested. VF and VUW1 were chosen by producing a worse quantization with a smaller threshold variation. VC was chosen because there is a cruise ship on the horizon similar to the target. The tested trackers used 3 input RAM nodes (3-bit bus address) because if performed best at previous simulations.

Table 3. Simulation S2: Mean tracking time ET (ms) x Frame that occurred the first tracking error.

| N | $T_5$ | $T_6$ | $T_7$ | $M_5$ | $M_6$ | $M_7$ |
|---|---|---|---|---|---|---|
| VF | 82 | 95 | 93 | 3 | 21 | X |
| $VB_1$ | 242 | 290 | 287 | X | X | X |
| $VB_2$ | 220 | 275 | 259 | X | X | X |
| VZ | 267 | 318 | 326 | X | X | X |
| VC | 105 | 107 | 107 | 19 | X | X |
| VS | 190 | 227 | 236 | 58 | X | X |
| $VP_1$ | 257 | 305 | 309 | X | X | X |
| $VP_2$ | 355 | 358 | 364 | X | X | X |
| $VUW_1$ | 130 | 146 | 152 | X | X | X |
| $VUW_2$ | 184 | 209 | 218 | 23 | 14 | X |
| VAE | 242 | 297 | 300 | X | X | X |

T5, T6 and T7: ET (ms) for tracking with GP5, GP6, and GP7 settings respectively; M5, M6, and M7: number of the first frame that the tracker fails with GP5, GP6, and GP7 settings respectively. GP5 has the same settings of simulation S1 (table 2). GP6 and GP7 have two discriminator layers: the first, GP, has 20 x 20 discriminators with sxp=5 and syp=5 and the second, GPD, denser than GP, has 10 x 10 discriminators with sxp=2 and syp=2. Both layers are centered on the target position provided by the predictor. GP5 and GP6 uses the segmentation method presented in section 3.2 and GP7 uses the improved segmentation method (this section).

Observing table 4, it is concluded that there is a relationship between the discriminator distributions and robustness against quantization errors. Comparing GP8 to GP9 and GP10 to GP11, it is clear that the tracker followed the target by more frames using discriminators spaced 1 xp and yp pixel than 2 pixels, although the first use less discriminators (25%). GP11 was the most efficient, surpassing the other in 8 of 9 tests.

### 3.3.5 Tracker with two parallel WiSARD neural networks

The WiSARD recognizes best a pre trained pattern (greater response) when pixels PO belonging to the object are quantized with a value and pixels PNO that do not belong to the object are quantized with another. If pixels are wrongly quantized, the network will respond with a smaller sum. In most segmentation methods, the probability of a PNO pixel being wrongly quantized is greater than that of a PO pixel, because main aim of these methods is to include PO pixels in a region RG to meet their requirements RQ. The segmentation focus is the inclusion of pixels PO in RG. The exclusion of PNO pixels from RG is merely a consequence of not respecting RQ. Excluding a PO pixel from RG is worse than adding a PNO pixel in RG. We conclude that there is a greater probability of occurrence of PNO pixels quantization errors than of PO pixels. The quantization proposed in this paper includes PO pixels in RG if their color belongs to an interval defined by thresholds L1 to L6. The PNO pixels exclusion from RG is not the priority. The exclusion only occurs if the PNO pixels are sufficiently different from PO pixels.

Designing a network where PO pixels have more weight in response entail more hits. The template matching algorithm of some Kernel based object trackers [20] associates weights to the SLW pixels. Pixels next to the SLW border have lower weights. The weighting increases the robustness of the matching since the peripheral pixels are the least reliable, being often affected by occlusions or clutter. Read [20] to understand the mathematical details.

The WiSARD, by definition, has no weights. To circumvent this problem, two parallel networks with different RAM node sizes at their discriminators can be used. Each discriminator response of the parallel network is an add of two discriminators responses working alone (3). The discriminators of each network cover a disjoint region and the union of both regions forms the region covered by the parallel discriminator. A discriminator with 3 input RAM nodes generates a higher response than a discriminator with 15 input RAM nodes by having a greater number of nodes. The first discriminator, which has more nodes (3 input RAM nodes), receives as input the quantized pixels inside the central part of the SLW (probably where the target is) and other discriminator, which has fewer nodes (15 input RAM nodes) receives as input the quantized pixels inside the peripheral part of SLW (probably where background pixels are)(figure 7). The word Parallel in PWOT comes from the innovative way of placing two discriminators in parallel to improve the WiSARD performance when quantization errors occur.

$$PDR = DR1 + DR2 \qquad (3)$$

Equation (3): PDR: parallel discriminator response; DR1: discriminator with 3 input RAM nodes response; DR2: discriminator with 15 input RAM nodes response.

By varying the quantity of pixels wrongly quantized, 129 simulations were performed to find the pixel percentage P of SLW covered by the discriminator with 3 input RAM nodes (pixels inside the blue region) and the (1-P) pixel percentage (pixels inside the pumpkin region) covered by the other one (with 15 input RAM nodes) to increase robustness against quantization errors (figure 7). For these tests we used the GP11 discriminator setting because it is the most effective against quantization errors (section 3.3.4).

The simulations showed that there is always a pixel percentage P (or percentage range) covered by the discriminator with 3 input RAM nodes which improves the performance when quantization errors occur. Using the optimal value of P, two WiSARD working in parallel is always more efficient than one network having 3 or 15 input RAM nodes working alone. Optimal value of P is not equal to all videos. At VF, P is between 40% and 45%. At VC, P is between 80% and 85%. At VUW1, P is between 17% and 22%. The optimal value depends on factors such as target and SLW sizes and target-background contrast. The online calculation of P can be done observing the decrease of the confidence level C or when the predictor perceives a failure.

Table 4. Percentage rise of wrongly quantized SLW pixels x Frame that occurred the first tracking error.

| N | MB | M8 | M9 | M10 | M11 |
|---|-----|-----|-----|-----|-----|
| VF | 34% | 46 | X | 21 | X |
| VF | 43% | 33 | 52 | 31 | 34 |
| VF | 65% | 15 | 15 | 15 | 18 |
| VC | 91% | X | X | X | X |
| VC | 93% | X | X | X | X |
| VC | 97% | 19 | 53 | 19 | 53 |
| VUW$_1$ | 42% | 72 | 73 | 31 | 73 |
| VUW$_1$ | 69% | 31 | 71 | 13 | 73 |
| VUW$_1$ | 122% | 14 | 20 | 15 | 20 |

MB: Relative percentage rise of wrongly quantized pixels; M8, M9, M10 and M11: first frame that a failure occurred using GP8, GP9, GP10 and GP11 settings respectively. GP8 comprises GP and GPD1. GP9 comprises GP and GPD2. GP10 comprises GP, GPD3, GPD4, GPD5 and GPD6. GP11 comprises GP, GPD7, GPD8, GPD9 and GPD10. GP has 20 x 20 discriminators with sxp=5 and syp=5. GPD1, GPD3, GPD4, GPD5 and GPD6 have 10 x 10 discriminators with sxp=2 and syp=2. GPD2, GPD7, GPD8, GPD9 and GPD10 have 5 x 5 discriminators with sxp=1 and syp=1. The layers GP, GPD1, GPD2, GPD3 and GPD7 are centered on the target position provided by the predictor. GPD4 and GPD8 are centered at the position of the greatest discriminator response in GP; GPD5 and GPD9 are centered at the position of second largest discriminator response in GP; GPD6 and GPD10 are centered at the position of third largest discriminator response in GP.

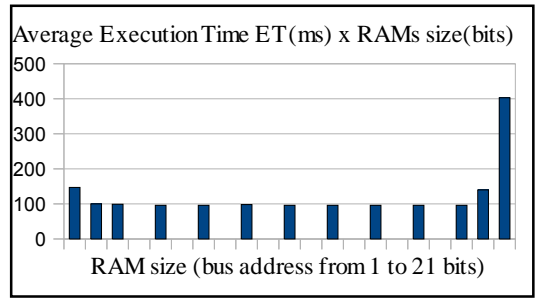

figure 6. Average execution time ET (ms) x RAM nodes sizes (bits) for VF video.

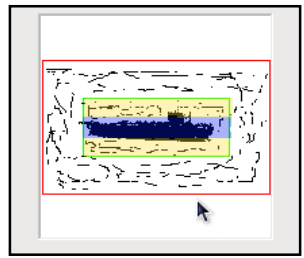

figure 7. The proposed parallel WiSARD discriminator has RAM nodes of two different sizes. It's reponse is an sum of two discriminators responses: the discriminator with more nodes receives as input the quantized pixels inside the central part of the SLW (blue region), the one with fewer nodes receives as input the quantized pixels inside the peripheral part of the SLW (pumpkin region). The green rectangle represents the SLW and the red one represents the ROI.

## 4. CONCLUSIONS

The results obtained by the simulations were better than expected. The computer used has little processing power, but the PWOT runs in real time, proving that the WNN WiSARD can be used for tracking. If a fewer quantity of RAM neurons and the best quantization (section 2.3.2) were used, the mean tracking time (ET) could be reduced in about 10 times or more without tracking failures.

The simulations in section 3.3.1 shows that the greater the number of WiSARD discriminators, the greater is ET, the greater is the ROI size and the lower is the probability to lose a fast target. When fewer quantization errors occur, the tracker commits fewer failures. The introduction of a Kalman position estimator improves performance without increasing considerably ET and it allows the tracker to check a tracking failure during the target detection phase.

A hybrid and innovative quantization method is described in section 3.3.2. When the target and background pixels are correctly quantized and increasing the number of discriminators around the position returned by the position estimator, the tracker follows the target by more frames without failure.

The tests performed in section 3.3.3 evaluated the changes in tracking performance when different bus address size nodes are used. The tests showed that bus address sizes between 2 and 14-bit have similar ET and are more efficient than other bus address sizes, though, 3 input RAM nodes compensate quantization errors more efficiently. An important conclusion is that the RAM node bus address size influences the robustness against quantization errors.

There is a relationship between the discriminator positions and robustness against quantization errors (section 3.3.4). Using fewer discriminators and less spacing in pixels among them, it permits a tracking with no failure by more frames.

An innovative way of using two WiSARD in parallel to increase the robustness against quantization errors was presented in section 3.3.5. The parallelism occurs at discriminator level (figure 7). Tracking with two parallel WiSARD is more efficient than using one because it can track the target by more frames when quantization errors occur. Thus, the parallel WiSARD can compensate the errors generated by the segmenter. The simulations showed that there is a percentage P of pixels covered by the WiSARD with 3 input RAM nodes and a percentage 1-P of pixels covered by the WiSARD with 15 input RAM nodes which causes an improvement of PWOT performance. P is not the same for all videos. The optimum value of P depends on factors like target size, SLW size and target contrast against the background. The online calculation of P can be done when the confidence level C is below a threshold or when the predictor points to a position that is impossible to the target be.

To improve the PWOT performance, it can be investigated new ways to distribute the discriminators, other kinds of predictors, new ways to organize two parallel neural networks, other kinds of features to train the network, methods to reduce the amount of features and the use of more than two WiSARD in parallel.

## ACKNOWLEDGEMENTS

My highest gratitude goes to the IpqM (Brazilian Naval Research Institute) and Civil Engineering Program – COPPE/UFRJ for all the help and support in carrying out this research.

## Authors

**Rodrigo S Moreira** is a reseacher at Brasilian Naval Research Institute and a D.Sc student at COPPE/Federal University of Rio de Janeiro. He has developing missile launch systems since 2007 and target tracking systems since 2013. He has published 1 article at BRICS international congress in 2013. His research focuses on video target tracking.

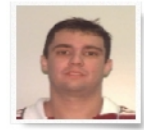

**Nelson F. F. Ebecken** is a Professor of Computational Systems at COPPE/Federal University of Rio de Janeiro. He is senior member of the IEEE and ACM. His research focuses on natural computing methodologies for modeling and extracting knowledge from data and their application across different disciplines. He develops models for complex systems, big data and integrates ideas and computational tools. In 2005 he was awarded as member of the Brazilian Academy of Sciences.

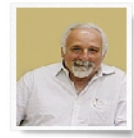